\documentclass[preprint,12pt]{elsarticle}

\usepackage{amssymb}
\usepackage{amsmath}
\usepackage{float}
\usepackage{xr}
\usepackage[hidelinks]{hyperref}
\journal{Robotics and Autonomous Systems}

\begin{document}

\externaldocument[supp-]{supplementary_material}

\begin{frontmatter}

\title{Learning a Resolution-Consistent Jacobian Field for Bio-Inspired Rigid-Soft Finger}

\author[1,2]{Tianyou Liang}
\ead{liangtianyou@nudt.edu.cn} 

\author[1,2]{Haisen Zeng}
\ead{zenghaisen@nudt.edu.cn} 

\author[1,2]{Shanjun Chen}
\ead{chenshanjun@nudt.edu.cn} 

\author[1,2]{YiMing Zhu}
\ead{yiming.zhu@nudt.edu.cn} 

\author[1,2]{Zhongyue Lu}
\ead{luzhongyue@nudt.edu.cn} 

\author[1,2]{Zirong Luo\corref{cor1}}
\ead{Luozirong@nudt.edu.cn}

\affiliation[1]{organization={College of Intelligence Science and Technology, National University of Defense Technology},
                addressline={},
                city={Changsha},
                postcode={410073},
                country={China}}

\affiliation[2]{organization={National Key Laboratory of Equipment State Sensing and Smart Support, National University of Defense Technology},
                addressline={},
                city={Changsha},
                postcode={410073},
                country={China}}

\cortext[cor1]{Corresponding author.}

\begin{abstract}
Bio-inspired tendon-driven rigid-soft coupled dexterous fingers exhibit strong nonlinearity 
and configuration-dependent sensitivity, making accurate modeling challenging. In discrete-time 
control, Jacobian-based kinematic algorithms typically rely on point-wise local linear approximations, 
which makes their performance sensitive to sensor sampling frequency and controller update frequency. 
To address this issue, we propose Jacobian Flow Matching (JFM), a structured learning framework based on Conditional 
Flow Matching (CFM), to learn a resolution-consistent Jacobian field that models actuation-to-motion 
transitions as a dynamical flow. The proposed framework supports both single-step prediction and continuous 
rollout via ODE integration, enabling consistent inference across temporal resolutions. Experiments on a 
tendon-driven rigid-soft finger show that the proposed method suppresses outlier errors and improves
 single-step prediction accuracy, reducing the global average RMSE by over 53\% compared with a baseline 
 discrete Jacobian learning approach. For long-horizon prediction, trajectories recovered via ODE integration 
 achieve higher fidelity under sparse sampling (Stride = 8), reducing the RMSE median by 14.43\% and the error 
 variance by 24.87\%. These results demonstrate that the learned  flow-based Jacobian field provides an effective 
 local model for offline multi-step trajectory optimization in rigid-soft coupled nonlinear systems.
\end{abstract}


\end{frontmatter}

\section{Introduction}
\label{sec1}

Rigid-soft coupled dexterous fingers promise versatile manipulation in cluttered and constrained 
settings~\cite{RN1, RN2}, yet reliable precision control remains difficult in practice. Unlike rigid-link 
manipulators, bio-inspired underactuated fingers exhibit strong actuation coupling and compliance-induced 
nonlinearities and hysteresis, so the mapping from actuation inputs to fingertip motion can change sharply 
with configuration and operating conditions (e.g. tendon routing)~\cite{RN3}.

Conventional approaches largely fall into two categories. Model-based methods derive kinematic or 
quasi-static mappings from finger design and calibrated parameters, and then apply feedback control 
using analytical Jacobians or locally linearized models~\cite{RN4, RN5}. While attractive for their interpretability, 
analytic transmission models are hard to maintain for tendon-driven soft underactuated fingers: tendon-sheath 
friction exhibits pronounced hysteresis and is often modeled under simplifying assumptions, which can limit 
accuracy and generalization across configurations and operating conditions~\cite{RN6, RN7}. In practice, this often 
necessitates repeated identification or retuning when hardware routing, load, or wear changes~\cite{RN8}.

However, learning-based control still faces a train-test gap under practical deployment. First, most data-driven models 
are tightly coupled to specific input conditions—whether step size, sampling rate, or command magnitude—and their 
performance degrades significantly when these conditions change. Second, the actuation-motion coupling of tendon-driven 
rigid-soft fingers varies markedly with configuration and operating conditions~\cite{RN10}, yet prevailing approaches 
focus on learning a direct pointwise Jacobian function while overlooking the holistic structure of the underlying 
Jacobian field—making them brittle under finite command increments and sparse sensing.

What is needed is not merely a pointwise local map, but a \textbf{resolution-consistent Jacobian field}: 
a Jacobian field that varies smoothly over the configuration space and remains consistent across sensing 
and control resolutions, so that its integrated predictions remain stable as the sampling interval changes. 
This shifts the core technical challenge from improving pointwise prediction accuracy alone to enforcing 
field-level consistency under deployment resolution changes.

To address this issue, we propose Jacobian Flow Matching (JFM), a Conditional Flow Matching (CFM)-based 
framework for learning resolution-consistent Jacobian fields. Our main contributions are as follows:

\begin{enumerate}
\item \textbf{Resolution-consistent Jacobian field for rigid-soft finger kinematics:} We formulate the actuation-motion 
relationship of rigid-soft coupled fingers as a smooth, configuration-dependent Jacobian field rather than a pointwise 
local map. The learned field is designed to remain consistent across sensing and control resolutions, improving motion prediction under discrete sampling.

\item \textbf{JFM learning of within-step Jacobian evolution:} We introduce a JFM training scheme that supervises 
the within-step evolution of the Jacobian field using adjacent observations. By normalizing 
the step interval and enforcing intermediate-time consistency, the proposed model supports both pointwise prediction and 
continuous-time rollout inference within a unified framework.

\item \textbf{Evaluation under sparse sampling:} We benchmark prediction and rollout fidelity across strides and horizons, 
and further present open-loop trajectory optimization to show that the learned field provides a useful local model for multi-step 
trajectory optimization.
\end{enumerate}

\section{Related Work}
\label{sec2}

Modeling soft bio-inspired robots remains an important research direction for next-generation robotic systems. Early approaches 
include piecewise constant-curvature (PCC) models \cite{RN11}, piecewise-smooth curvature parameterizations \cite{RN12}, Cosserat-rod theory \cite{RN13}, 
and finite-element methods (FEM) \cite{RN14}. These methods provide useful physical interpretations, but for tendon-driven soft dexterous fingers, 
routing-dependent friction, backlash-induced hysteresis, configuration-dependent coupling, and elastic recovery make purely analytic models 
difficult to use for reliable control \cite{RN15, RN16}.

To bridge the gap between analytic modeling and fully end-to-end policies, recent work has increasingly explored learning-based structured local models for control. 
Representative examples include actuation-to-motion mappings that reduce reliance on explicit transmission modeling \cite{RN10}, and configuration-dependent Jacobians 
or locally linear dynamics learned from data for optimization-based control \cite{RN17}. As surveyed by Chen et al. \cite{RN25}, Jacobian-based methods for soft robots 
rely on online updating from actuation-state data, yet their oversimplified linearization struggles with complex nonlinearities. More broadly, multimodal proprioceptive 
sensing has enabled structured local models that remain compatible with control, such as Jacobians inferred from enriched sensory observations \cite{RN19}.

Existing approaches face three specific limitations. First, end-to-end reinforcement learning \cite{RN10} maps motor currents directly to end-effector positions without 
an explicit process model, leading to poor generalization when sampling rates or command magnitudes change. Second, neural Jacobian estimators \cite{RN17} learn a pointwise 
map \(J(s_t)\) without constraints on intermediate states, causing performance degradation under ODE integration. Third, visual-based methods \cite{RN18,RN19} rely on 
external cameras for closed-loop correction or Jacobian inference, while occlusion and real-time computational overhead make such approaches unnatural and inefficient 
for dexterous multi-finger hand systems; moreover, the visuomotor Jacobian field\cite{RN18} learns only a pointwise map from images, lacking consistency along continuous 
trajectories and across varying step sizes.

Despite the above efforts, most learned Jacobian models remain pointwise functions rather than true fields: they map a sampled state to a fixed local linearization, 
without enforcing consistency across neighboring states or along continuous trajectories. For tendon-driven rigid-soft fingers, where sensitivity varies smoothly 
with configuration and actuation history, such pointwise approximations are inherently insufficient—the effective Jacobian at an intermediate state along a 
finite-step transition cannot be inferred from the endpoints alone. This motivates learning a resolution-consistent Jacobian field that captures the continuous 
evolution of local sensitivity.

In this paper, we address this issue by learning a resolution-consistent Jacobian field from multimodal proprioception, targeting consistent finite-step 
motion prediction across sensing and control scales.

\section{Proposed Method}
\label{sec3}
\subsection{Finger Platform Overview}
\begin{figure}[htbp]
    \centering
    \includegraphics[width=1\textwidth, trim=0 90 0 0, clip]{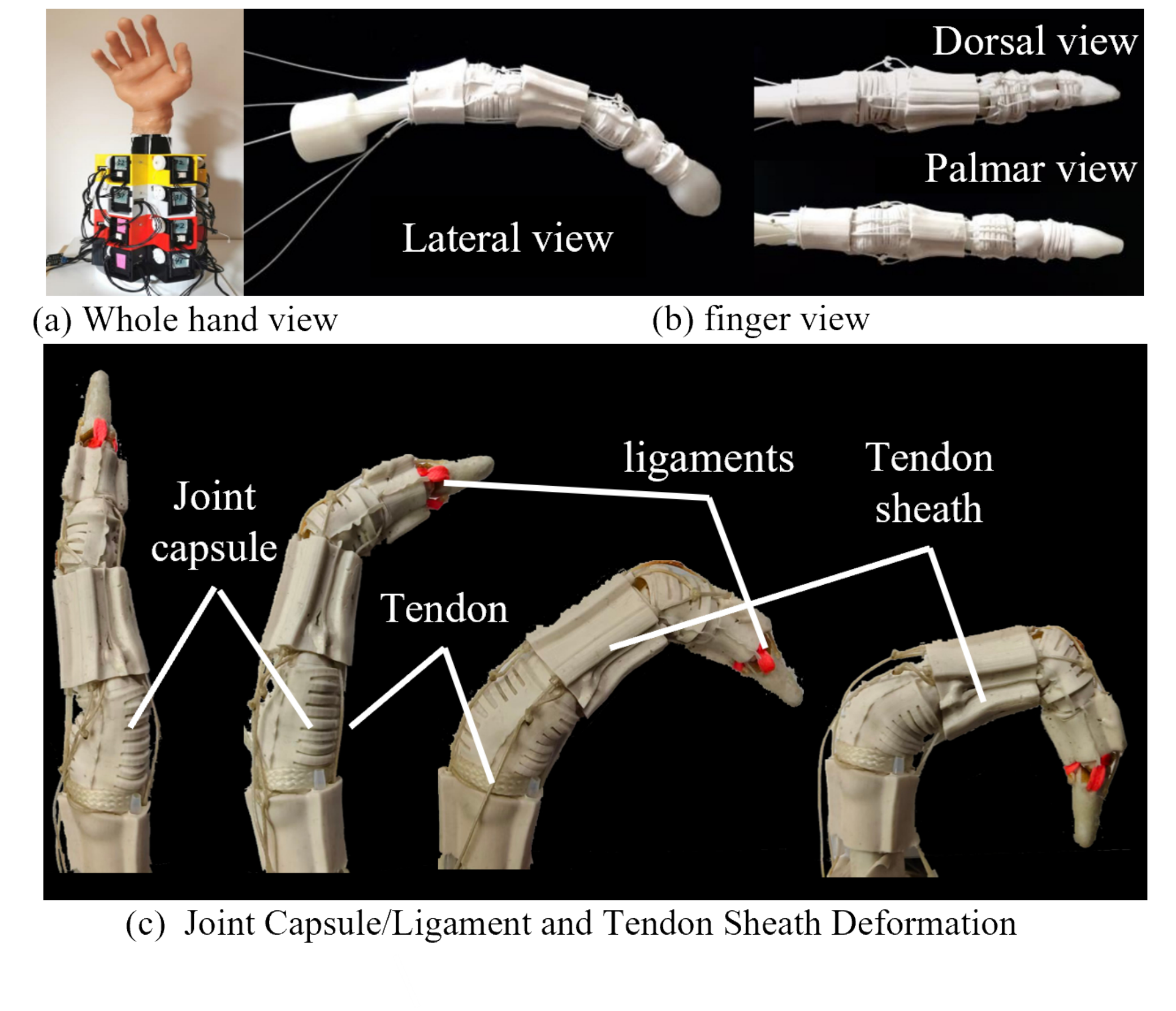}
    \caption{Bioinspired anthropomorphic robotic finger \cite{RN15,RN16}}
    \label{fig:finger}
\end{figure}

As illustrated in Fig.~\ref{fig:finger}, the bio-inspired finger under investigation is designed as a rigid-flexible coupled system\cite{RN15, RN16}. By replicating the key biomechanical 
structures of the human finger, this design achieves anthropomorphic passive advantages—such as structural stability, physical compliance, and functional coupling—while 
remaining compatible with tendon-driven actuation and manufacturing constraints. Specifically, the finger integrates and simplifies these features into three functional 
layers: (i) a rigid skeletal layer, composed of 3D-printed phalanges; (ii) a flexible joint capsule and ligament layer, which includes PTFE-coated articular interfaces 
and a silicone joint capsule embedded with woven fiber ligaments; and (iii) a tendon transmission layer. The extensor mechanism forms a complex transmission network that 
achieves multi-degree-of-freedom motion through coordinated pulling of multiple tendons.

As illustrated in Fig.~\ref{fig:finger}(c) during finger flexion, although this system exhibits good physical compliance, its inherent material and structural characteristics pose 
significant challenges for accurate modeling. The main sources of nonlinearity and variability in our finger include: (i) Configuration-dependent and time-varying behavior 
induced by structure and materials: Due to the combined action of the flexible joint capsule/ligament structure and the tendon transmission network, 
the system exhibits hyperelastic and viscoelastic responses during stretching, compression, and contact, accompanied by effects such as friction, backlash, 
and hysteresis. As a result, the same servo position change $\Delta q_t$ may produce significantly different motion increments $\Delta x_t$ under different phalanx 
orientation $x_t$, which makes it hard to capture the behavior over the workspace with a single globally valid analytical model. (ii) Scale-dependent sensitivity under discrete control. 
Because the effective sensitivity depends on step size, a pointwise local linearization may be inaccurate for finite command increments. A typical remedy is ODE-based rollout, which 
subdivides the step and integrates incremental predictions. But this remedy shifts the requirement: the learned Jacobian must now function as a consistent vector field, where the 
integrated result over sub-steps agrees with the observed finite displacement. Most existing pointwise models, trained without intra-step supervision, fail to satisfy this. 
(iii) Inter-frame evolution under limited observation frequency: Since observations are discretely sampled, the Jacobian 
field may change between adjacent sampling frames. If one only learns/uses point mappings at sampling frames, it is easy to overlook the continuous evolution laws between frames, 
thereby affecting the stability of rollout predictions, or weakening interpolation/extrapolation capabilities when frames are dropped.

We do not explicitly identify the constitutive parameters of the flexible materials, nor do we assume they are strictly constant. This paper only makes the following weak 
assumption: within the time window of a single trial, the transmission and material characteristics change relatively slowly, so that there exists a learnable equivalent 
local input-output relationship conditioned on the current configuration and actuation state. Unmodeled effects such as hysteresis and friction are treated as disturbances 
that are absorbed implicitly by the learned model within the training distribution. Accordingly, our goal is to learn a continuous Jacobian field that improves step-size 
consistency of motion prediction across different sampling/control strides, instead of relying on Jacobian mappings tied to a fixed discrete step.

\subsection{Resolution-Consistent Jacobian Field}
\label{sec:3_2}

Let \(x \in \mathbb{R}^d\) denote the task-space kinematic state and \(q \in \mathbb{R}^m\) the servo positions.
In discrete-time control, a finite increment \(\Delta q_t\) produces an observed one-step change
\(\Delta x_t := x_{t+1} - x_t\).

This discrete transition is modeled using a local Jacobian field \(J(s_t,\Delta q_t)\) that satisfies
\begin{equation}
\Delta x_t \approx J(s_t, q_t)\,\Delta q_t 
\label{eq:res_jac}
\end{equation}
where $J(s_t,\Delta q_t)\approx{\frac{\partial x}{\partial q}\mid}_{s_t}$ denotes the local actuator-to-task sensitivity, parameterized by the observation feature $s_t$(proprioception, 
actuator states, and short history) as a compact descriptor of the current finger configuration.
Once learned, \(J\) should be used in two ways: (i) a single-step pointwise prediction via \eqref{eq:res_jac}, 
or (ii) an ODE-based rollout that subdivides the step and integrates predictions over intermediate states. 
If the field is well-behaved, the latter can be more accurate because it exploits continuous-time information.

However, prior pointwise training approaches (e.g., \cite{RN5,RN17,RN18}) exhibit two limitations. First, when ODE 
integration is applied to their learned field, accuracy may degrade compared to pointwise prediction 
(see Section~\ref{sec:4_1}, Fig.~\ref{fig:JFM_Baseline}). Second, for the same reason,when the command magnitude changes (e.g., under sparse sampling 
with larger strides), the model may suffer from performance degradation. These observations suggest that 
the learned field may not be well-suited for continuous integration and may lack robustness to step-size 
variations.

A key reason is that most existing methods learn pointwise Jacobian \textit{mappings}, not a true Jacobian \textit{field}. They are trained only on 
endpoint supervision, with no constraints on intermediate states \(x_\tau\) (\(\tau \in (0,1)\)). As a result, the model may behave arbitrarily at 
unobserved states. In applications that inevitably query these intermediate states—e.g., ODE integration, sparse-sampling interpolation, or multi-step 
rollout—unreliable Jacobian estimates can accumulate and gradually bias the control model, leading to significant downstream errors.

\subsection{Jacobian Flow Matching}
\begin{figure}[htbp]
    \centering
    \includegraphics[width=1\textwidth, trim=0 0 0 0, clip]{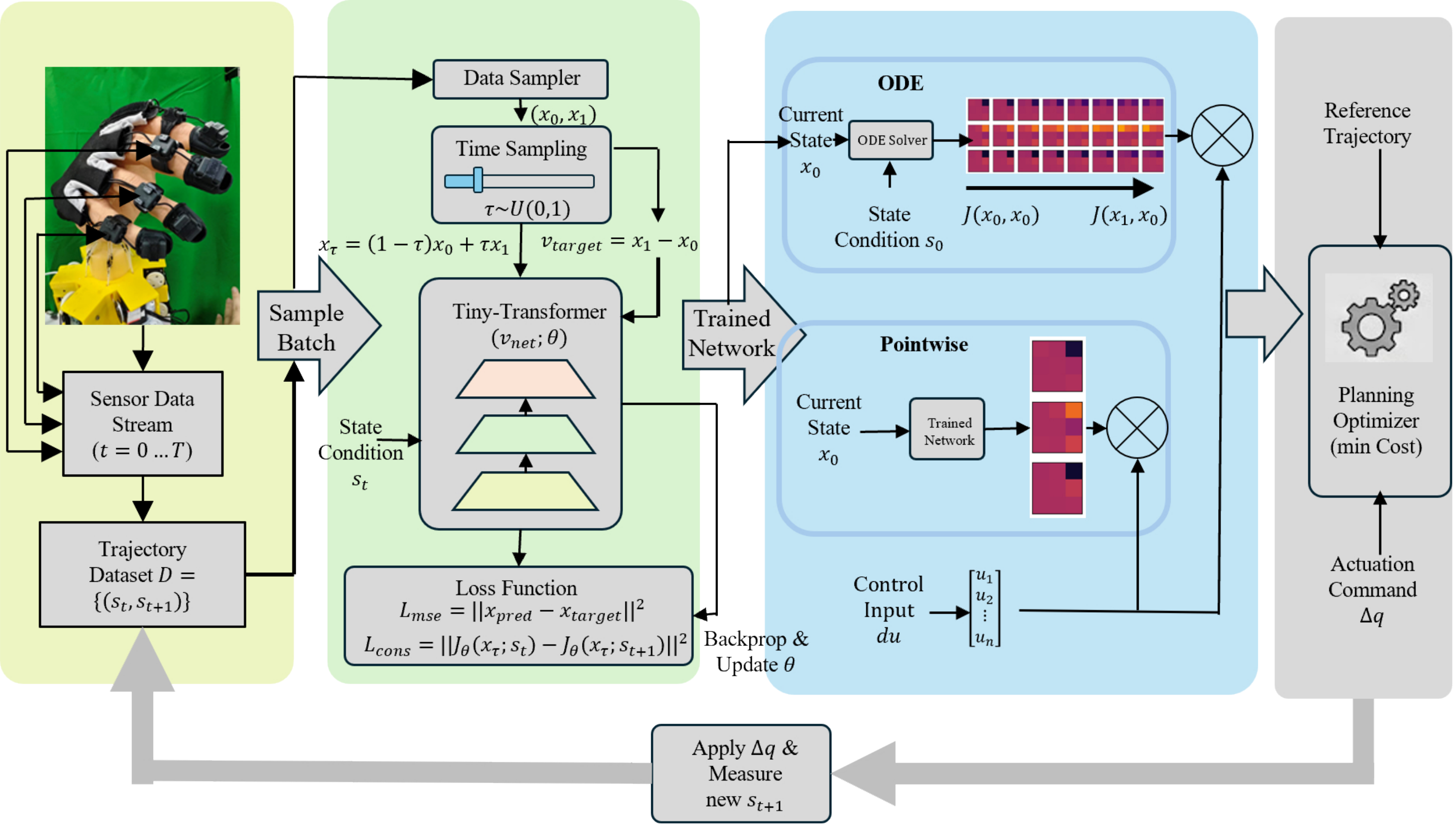}
    \caption{Overall framework of the proposed learning and inference pipeline}
    \label{fig:framework}
\end{figure}

We argue that resolution-consistent Jacobian learning should incorporate the within-interval evolution over each sampling period, instead of 
relying solely on the starting state. Concretely, we use the observed displacement $\Delta x_t$  over one interval as supervision, which captures 
the nonlinear cumulative effects produced while executing $\Delta q_t$. By normalizing each transition to a unit-time flow, the learning target 
becomes less sensitive to the absolute stride and remains consistent with the discrete integration process. This yields a single continuous 
Jacobian field that can be queried at intermediate times based on the inferred state evolution, rather than a collection of stride-specific Jacobian mappings.

In practice, we condition the Jacobian generator on the conditioning state $s_t = (x_t, q_t)$, which summarizes the instantaneous kinematic state and actuation 
configuration. This context provides the necessary intra-interval information (e.g., motion regime and actuation modality) to produce Jacobian estimates 
that are consistent with the motion that actually occurs over the sampling interval

Even with this conditioning, directly learning discrete step-to-step mappings, however, often struggles to enforce a coherent continuous-time evolution for rigid--soft coupled systems. 
To obtain a principled characterization of the integral evolution—from an instantaneous velocity description to finite-step displacements—we introduce 
a Jacobian-field learning framework based on CFM. 
Within this framework, we highlight two practical benefits. 
First, the CFM-style normalized path construction provides a convenient training parameterization that reduces explicit dependence on absolute step duration and enables dense within-step supervision. 
Second, CFM achieves implicit data augmentation through stochastic resampling of temporal evolution paths, which enables continuous-time interpolation between discrete observations when sampling is sparse. 
Importantly, we do not attempt to identify the true physical instantaneous Jacobian everywhere in continuous time. Instead, we learn a normalized within-step Jacobian field whose integral behavior is consistent 
with the observed finite displacement over a discrete control interval.

As illustrated in Fig.~\ref{fig:framework}, the framework consists of two modules:
\begin{enumerate}
    \item \textbf{Data-driven Jacobian field modeling:} We parameterize the Jacobian field using a lightweight Transformer network, conditioned on the 
    initial state \(s_t\) and the normalized time \(\tau \in [0,1]\). Training combines an MSE term with a consistency regularizer \(L_{cons}\) to encourage 
    coherent evolution along the time axis.

    \item \textbf{Flexible inference for prediction:} Because the learned field is continuous in time, it supports two inference modes. For dense 
    sampling, a single-step pointwise query provides fast prediction. Under sparse sampling, multi-step ODE integration leverages the continuous-time formulation 
    to maintain accuracy and reduce drift. Both modes share the same trained network.
\end{enumerate}

The velocity field is trained with a CFM-based endpoint-coupled path construction and a stochastic time-slice supervision mechanism. During inference, the learned velocity 
field can be used either as a pointwise Jacobian or integrated via an ODE solver for multi-step prediction.

For two adjacent observed states $(x_t, x_{t+1})$, we sample $\tau \sim U(0,1)$ and construct the intermediate state via a predefined generation path (e.g., 
the Optimal Transport (OT) path) as $x_\tau = \psi_\tau(x_t, x_{t+1})$. In this design, the velocity field is parameterized as a Jacobian-structured bridging field:

\begin{equation}
    v_\theta(x_\tau, x_t, q_t, \Delta q) = J_\theta(x_\tau, x_t, q_t) \Delta q \label{eq:linear_jacobian}
\end{equation}

Thus, we supervise the Jacobian field using the observed finite displacement. Rather than modeling the true continuous-time velocity, we learn a step-equivalent mapping that predicts the net displacement 
caused by the joint increment \(\Delta q_t\). Concretely, for any intermediate state \(x_\tau = \psi_\tau(x_t, x_{t+1})\) on a bridge path within the same control step, we require the predicted 
step displacement \(J_\theta(x_\tau)\Delta q_t\) to agree with the measured \(\Delta x_t\). This yields

\begin{equation}
    L_{\text{mse}}(\theta) = \mathbb{E}_{t, \tau \sim U(0,1)} \left\| \Delta x_t -J_\theta(\psi_\tau(x_t, x_{t+1}), x_t, q_t) \Delta q_t \right\|^2 \label{eq:L_mse_jacobian}
\end{equation}

Sampling \(x_\tau\) provides dense supervision along the step, which encourages resolution-consistent predictions when the step is subdivided during inference (e.g., 
under sparse sampling or ODE-based rollout).

However, a potential issue arises with the current loss formulation. CFM additionally introduces static conditioning inputs: the current state $x_t$ and actuation 
command $q_t$. Since each training instance consists of only two frames $(x_t, x_{t+1})$ as CFM endpoints, minimizing the loss function in Equation \ref{eq:L_mse_jacobian} may lead 
to a shortcut: the model could overly rely on the static conditions $(x_t, q_t)$ to explain $\Delta x_t$, while failing to fully leverage the 
randomly sampled intra-step states $x_\tau$.

To mitigate this issue, we introduce a dual-view consistency condition. Given a triplet $(x_t, x_\tau, x_{t+1})$, we assume that the motion over 
a single sampling interval is smooth and locally well-approximated in a neighborhood of the intermediate state $x_\tau$. Under this assumption, 
a physically consistent local model should yield comparable instantaneous sensitivities at the same intermediate state, regardless of whether 
the transition is parameterized from the start state $s_t = (x_t, q_t)$ or from the end state $s_{t+1} = (x_{t+1}, q_{t+1})$. Intuitively, 
the local actuation-motion sensitivity at $x_\tau$ should reflect the realized motion over the interval, rather than the particular endpoint used for conditioning.

Admittedly, hysteresis may break exact consistency between the two views. Over short monotonic steps, however, enforcing such consistency serves as a reasonable 
regularization to learn a smooth Jacobian field.

Accordingly, we impose the dual-view consistency constraint: for any sampled intermediate state $x_\tau$, the Jacobians predicted under the two views should be close:

\begin{equation}
    L_{\text{cons}}(\theta) = \mathbb{E}_{t, \tau \sim U(0,1)} \left\| J_\theta(x_\tau; s_t) - J_\theta(x_\tau; s_{t+1}) \right\|^2 \label{eq:L_cons_jacobian}
\end{equation}

This regularization term forces the model to produce consistent local linearizations for the same intra-step state under different endpoint-based conditions, thereby 
suppressing reliance on static conditioning and encouraging the model to learn a truly state-dependent Jacobian field. The final training objective is:

\begin{equation}
    L_{\theta} = L_{\text{mse}}(\theta) + \lambda_{\text{cons}} L_{\text{cons}}(\theta) \label{eq:L_total_jacobian}
\end{equation}

where $\lambda_{\text{cons}}$ is a hyperparameter that balances the weight of the consistency term.

\subsection{Bidirectional validation of the learned Jacobian field}

We validate the learned Jacobian field from two complementary perspectives: \textbf{forward} (predicting motion given actuation) and \textbf{inverse} 
(recovering actuation given a desired target). Both validations are conducted offline and do not claim real-time closed-loop performance.

\paragraph{Forward validation (open-loop rollout).}
Given an actuation command sequence $\{\Delta q_{t+k}\}_{k=0}^{H-1}$, the learned field can predict the resulting task-space trajectory via two 
inference modes. The pointwise mode provides fast single-step prediction:

\begin{equation}
    x_{t+1|t} = x_t + J_\theta(s_t) \Delta q_t, \label{eq:forward_point}
\end{equation}

while the ODE mode integrates the continuous-time velocity representation for multi-step rollout under sparse sampling:

\begin{equation}
    x_{t+H|t} = x_t + \int_0^{H} J_\theta(x(\tau), s(\tau)) \, \Delta q_\tau \, d\tau,
    \label{eq:forward_ode}
\end{equation}
where $J_\theta$ is the pointwise Jacobian from the learned field.ODE-based 
rollout significantly outperforms pointwise prediction under sparse sampling (e.g., stride=8), demonstrating the field's continuous-time consistency.

\paragraph{Inverse validation (planning feasibility).}
To test whether the field can recover feasible actuation commands from desired targets, we formulate a single-step planning problem:

\begin{equation}
    \Delta q_t^* = \arg\min_{\Delta q_t \in \Delta \mathcal{Q}} \| x_t + J_\theta(s_t) \Delta q_t - x^* \|_Q^2 + \|\Delta q_t\|_R^2,
    \label{eq:inverse_opt}
\end{equation}

where $Q \succeq 0$ and $R \succ 0$ are weighting matrices, and $\Delta \mathcal{Q}$ encodes actuator limits. The optimization is performed 
without closed-loop feedback.

Together, the forward and inverse validations confirm that the learned Jacobian field is both accurate for prediction and compatible with optimization-based 
planning. Extending this framework to closed-loop, real-time control is left for future work.

\section{Experiment}
To comprehensively evaluate the proposed JFM framework, we conduct four sets of experiments. Section~\ref{sec:4_1} compares the proposed JFM against baseline methods under both 
pointwise and ODE-integrated inference modes. Section~\ref{sec:4_2} examines the resolution consistency of the learned Jacobian field by evaluating the same model under 
varying strides ($s = 1$, $4$, and $8$). Section~\ref{sec:4_3} ablates the dual-view consistency loss $\mathcal{L}_{\text{cons}}$ to verify its effectiveness in mitigating 
start-pose bias and improving mid-sequence modeling. Section~\ref{sec:4_4} validates the feasibility of using the learned Jacobian field for optimization-based planning 
through an open-loop inverse inference task. All evaluations use only the middle finger with in-plane flexion/extension, excluding out-of-plane and multi-finger motions to 
minimize interference from external variables such as inter-finger coupling. Generalization to spatial and multi-finger settings is left for future work.

 Unless otherwise specified, all experiments employ the TinyTransformer backbone; the ablation on backbone architecture is presented exclusively in Section~\ref{sec:4_1}. 
 Generalization to spatial and multi-finger settings is left for future work.

The data acquisition platform collects data via ROS2, with the rigid-soft coupled dexterous hand equipped with an instrumented 
glove (Fig. \ref{fig:framework}). State observations are the Euler angle orientations of each phalanx. All training and inference are performed 
on a workstation with an NVIDIA RTX 6000 Ada GPU.

\subsection{Comparison with Baselines}
\label{sec:4_1} 
We evaluate prediction performance under different observation strides to simulate varying effective sampling frequencies. 
The comparison includes four training/inference combinations: JFM-ODE, JFM-Point, BASE-Point, and BASE-ODE. 
The BASE-Point model adapts the pointwise paradigm of \cite{RN18} to our proprioceptive setting
\footnote{In this experiment, the Point-Map baseline adapts the pointwise learning paradigm 
from \cite{RN18} to our proprioceptive setting, replacing visual optical flow with phalangeal IMU measurements.}, 
while BASE-ODE serves as a negative control to assess whether ODE rollout alone can compensate for a pointwise-trained model.

\begin{figure}[!ht]
    \centering
    \includegraphics[width=1\textwidth, trim=0 0 0 0, clip]{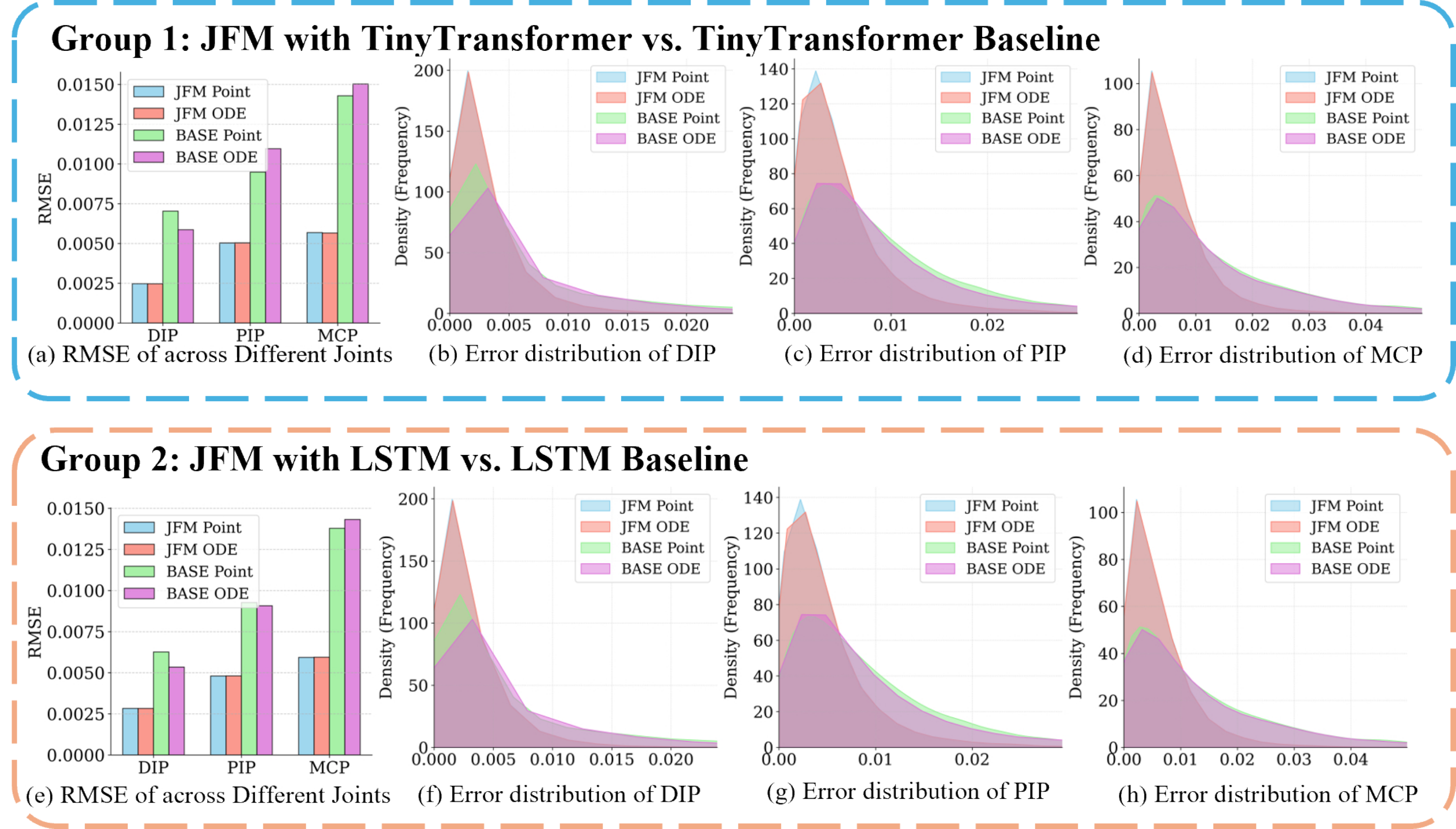}
    \caption{Servo-to-joint angle mapping (Z-axis flexion)}
    \label{fig:JFM_Baseline}
\end{figure}

To isolate the effect of the training paradigm, we evaluate all four training/inference combinations across two backbone architectures: 
TinyTransformer and LSTM (see the Supplementary Material for architectural details). The only difference within each backbone condition is how 
the model is trained and inferred. This controlled comparison ensures that any performance gain is attributed to the training framework, not 
to changes in network capacity or input preprocessing. Fig.~\ref{fig:JFM_Baseline} presents the results for both backbones, organized as Group~1 (TinyTransformer) and Group~2 (LSTM).

\begin{itemize}
    \item[\textbullet] \textbf{JFM-ODE}: Trained with our JFM framework, inferred via ODE integration.
    \item[\textbullet] \textbf{JFM-Point}: Trained with JFM, inferred via single-pass pointwise update.
    \item[\textbullet] \textbf{BASE-Point}: Conventional pointwise training and inference (adapted from \cite{RN18}).
    \item[\textbullet] \textbf{BASE-ODE}: Conventional pointwise training but with ODE integration forcibly applied (negative control).
\end{itemize}

As shown in Fig.~\ref{fig:JFM_Baseline}, the three joints correspond to the distal interphalangeal (DIP, Distal interphalangeal joint), proximal interphalangeal (PIP, Proximal interphalangeal joint), 
and metacarpophalangeal (MCP, Metacarpophalangeal joint) joints, respectively. Subfigures (a) and (e) report the root mean square error (RMSE) for each joint, where 
blue denotes pointwise prediction and orange denotes ordinary differential equation (ODE) integration. Subfigures (b)-(d) and (f)-(h) show the corresponding kernel density estimation (KDE) error distributions.

The proposed JFM achieves consistently lower errors across all three joints. For Group~1 (TinyTransformer backbone), JFM reduces 
the global average RMSE by 57.27\% compared to the baseline; for Group~2 (LSTM backbone), the reduction is 53.68\%.

The KDE distributions in Fig.~\ref{fig:JFM_Baseline} (b)-(d) further reveal the error characteristics for Group~1. JFM yields a 
tighter peak in the low-error region (\(\text{RMSE} < 0.005\)) for all three joints: the peak density ratios (JFM / baseline) 
are 1.80\(\times\) for the DIP joint, 1.50\(\times\) for the PIP joint, and 1.17\(\times\) for the MCP joint, averaging 1.49\(\times\) 
across the three joints. The baseline, in contrast, exhibits a heavy tail extending beyond 0.05. JFM successfully truncates this long tail. 
Taking an RMSE threshold of 0.015, the percentages of samples below this threshold are as follows: for the DIP joint, JFM achieves 99.6\% versus 
baseline's 88.2\%; for the PIP joint, 95.3\% versus 79.8\%; for the MCP joint, 93.8\% versus 65.9\%. Averaged across all three joints, 96.2\% of 
JFM samples fall within \(\text{RMSE} = 0.015\), compared to only 77.9\% for the baseline.

Subfigures (f)-(h) show similar results for Group~2 (LSTM backbone). The peak density ratios (JFM / baseline) are 1.48\(\times\) for DIP, 1.63\(\times\) 
for PIP, and 1.10\(\times\) for MCP, averaging 1.40\(\times\) across the three joints. For the RMSE \(< 0.015\) threshold: DIP achieves 99.1\% versus 88.3\%; 
PIP achieves 95.9\% versus 80.9\%; MCP achieves 93.2\% versus 67.0\%. Averaged across all three joints, 96.1\% of JFM samples fall below 0.015, compared to 78.8\% for the baseline.

Two consistent observations can be made from Fig.~\ref{fig:JFM_Baseline}. Taking Group~1 as an example, first, the gap between pointwise and ODE-based 
inference is small for JFM: the average difference across the DIP, PIP, and MCP joints is 0.22\% (with a maximum of 0.29\% at the MCP joint), indicating 
that the learned field is well aligned with continuous-time interpolation. Second, the baseline shows a larger performance drop under ODE integration: 
for the PIP joint, the RMSE increases by 14.8\% when switching from pointwise prediction to ODE integration. This is counter-intuitive: even under locally 
linear conditions, ODE integration degrades performance for pointwise-trained models, suggesting weaker integrability and reduced resolution consistency.

\subsection{Resolution Consistency Analysis}
\label{sec:4_2} 
To examine whether the learned Jacobian field is consistent across temporal resolutions, we evaluate the same model under different step sizes (strides). 
Specifically, for an observed transition \((x_t, q_t) \rightarrow x_{t+1}\), we compare (i) a single-step update using the pointwise Jacobian prediction 
and (ii) a multi-substep rollout where the velocity field is evaluated at intermediate states and integrated over the unit-time interval. If the learned 
field is resolution-consistent, subdividing the step should not change the net displacement, and the two rollouts should agree.

\begin{figure}[H]
    \centering
    \includegraphics[width=1\textwidth, trim=0 0 0 0, clip]{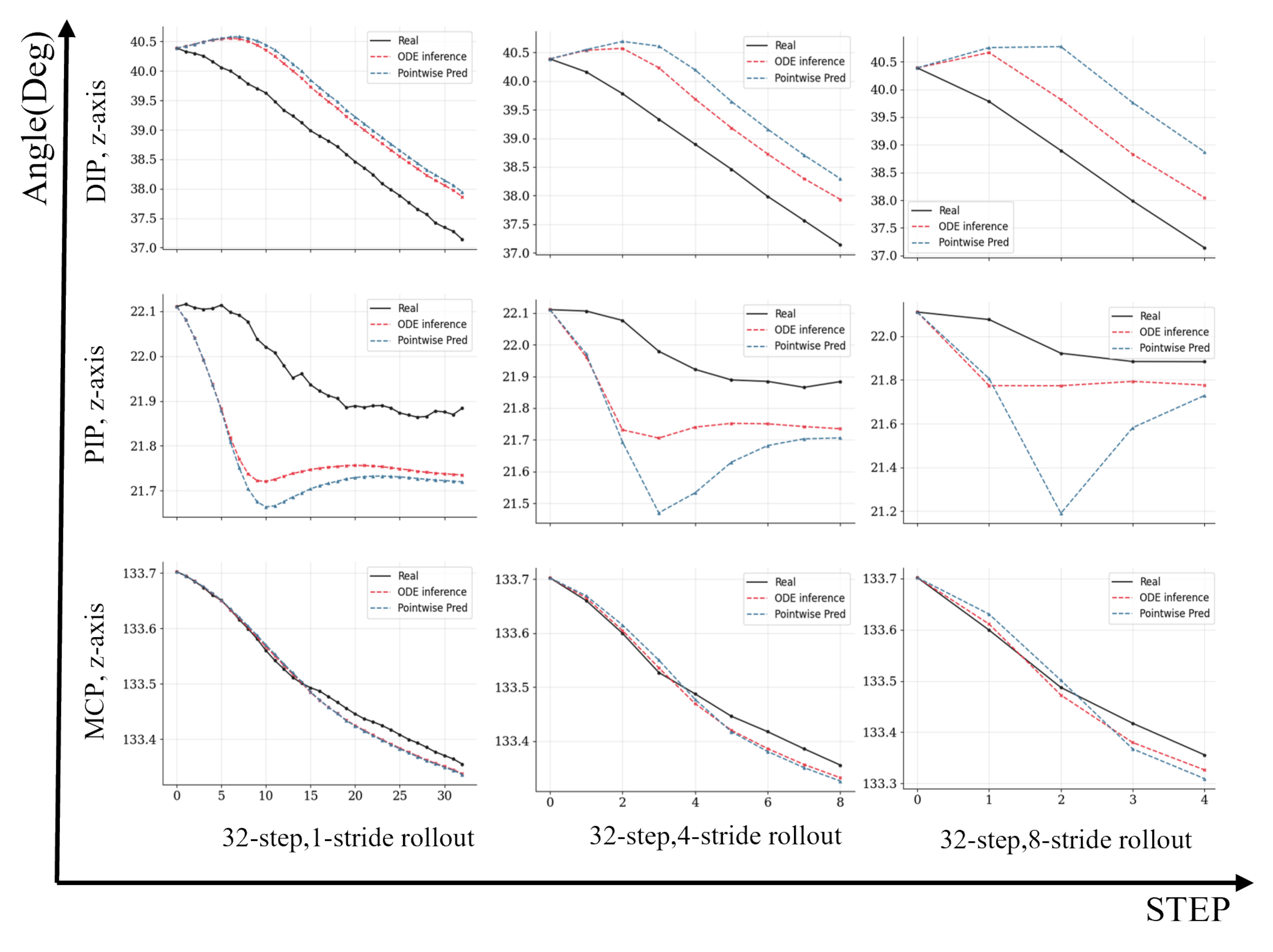}
    \caption{Trajectory prediction for different strides.}
    \label{fig:trajectory}
\end{figure}

Fig.~\ref{fig:trajectory} presents a 3$\times$3 grid of trajectory predictions. The columns correspond to the DIP, PIP, and MCP joints (distal interphalangeal, 
proximal interphalangeal, and metacarpophalangeal, respectively), and the rows correspond to strides 1, 4, and 8. Under dense sampling (stride=1), pointwise and ODE 
predictions nearly overlap across all joints, as expected under local linearity. As stride increases to 4 and 8, the pointwise predictions begin to drift, while the 
ODE-integrated rollouts remain stable. This suggests that the learned JFM field is substantially more resolution-consistent than pointwise baselines: its integrated 
predictions remain stable as the step is subdivided, especially under sparse sampling.

\begin{figure}[!ht]
    \centering
    \includegraphics[width=1\textwidth, trim=0 0 0 0, clip]{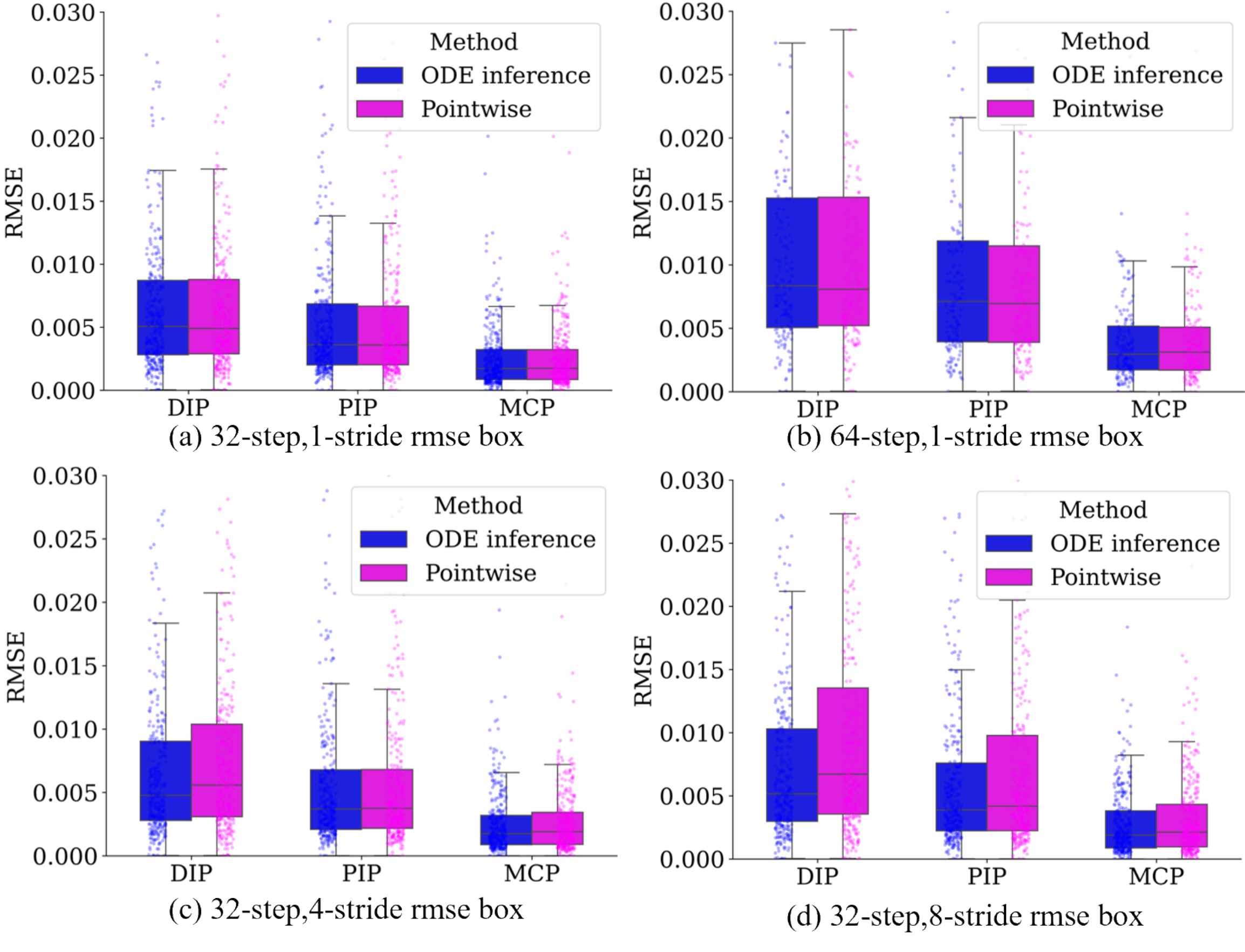}
    \caption{the open-loop trajectory prediction results' box}
    \label{fig:boxplot}
\end{figure}

Fig.~\ref{fig:boxplot} reports RMSE boxplots over 160 test trajectories, providing statistical evidence. At stride=1, both inference modes perform similarly. 
As stride increases to 4 and 8, the pointwise baseline degrades substantially, showing higher median error and wider interquartile range, while ODE inference 
maintains a compact error distribution. Quantitatively, at stride=8, ODE inference improves the RMSE median by 14.43\% and reduces variance by 24.87\%.

\subsection{Ablation on $L_{\text{cons}}$}
\label{sec:4_3} 
\begin{figure}[htbp]
    \centering
    \includegraphics[width=1\textwidth, trim=0 0 0 0, clip]{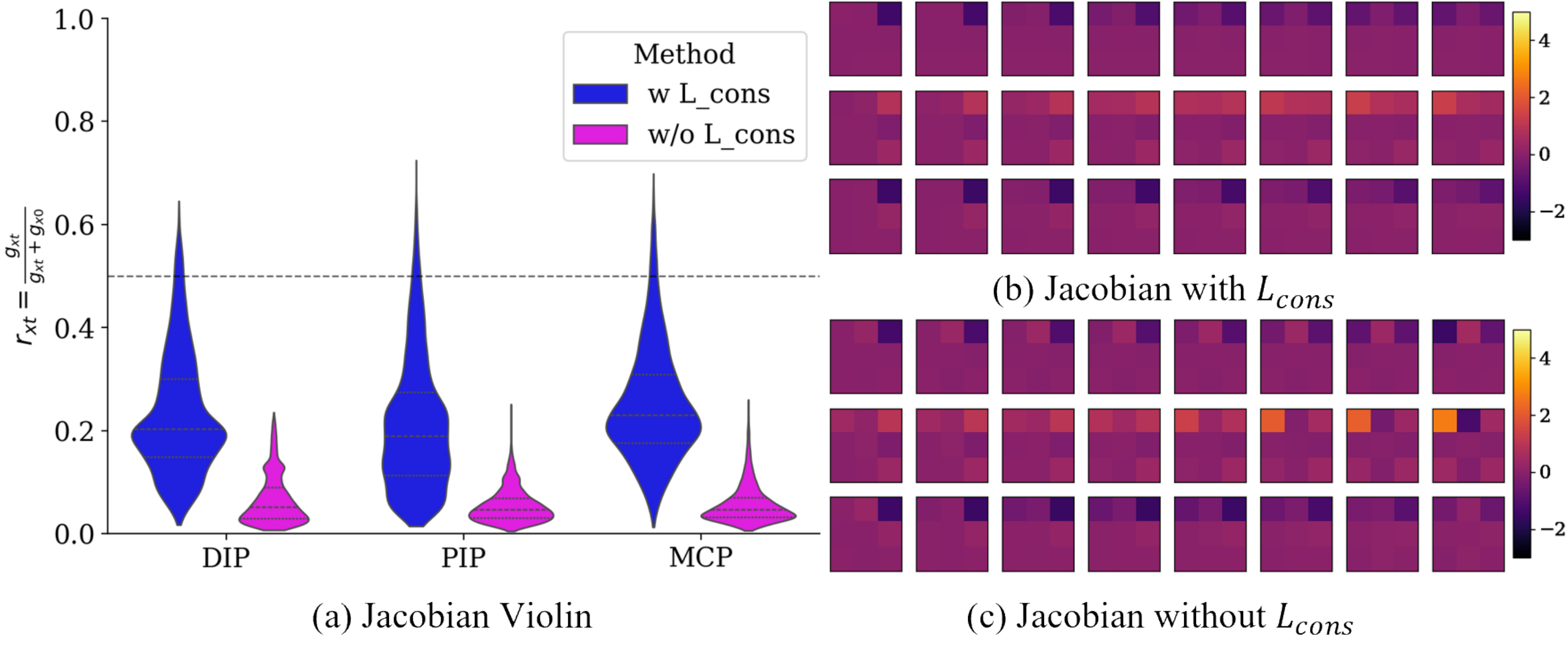}
    \caption{the open-loop trajectory prediction results' box}
    \label{fig:lcons}
\end{figure}

When conditioning only on the start pose \(x_0\) and start command \(q_0\), the model naturally acquires a "low-uncertainty, strongly constrained" anchor during training. 
Since the start frame and subsequent trajectories share the same generation pathway in loss backpropagation, the network tends to develop an opportunistic strategy: prioritizing 
local consistency near the start pose to reduce overall reconstruction error. However, for intermediate interpolated frames—which exhibit higher-frequency variations and are more 
sensitive to dynamics and contact—the model's fitting strength is relatively insufficient, manifesting as jitter, phase lag, or amplitude smoothing. The proposed dual-view consistency 
loss \(L_{cons}\) is designed to mitigate this issue. Its effectiveness is experimentally verified in Fig.~\ref{fig:lcons}.

The experimental results validate the effectiveness of this design, as shown in Fig.~\ref{fig:lcons}. The violin plot on the left 
compares the distribution differences of each joint under the normalized error metric with and without \(L_{\mathrm{cons}}\). On the 
vertical axis, \(g_{x_t}\) represents the corresponding Jacobian gradient magnitude for a given pose \(x_t\), as defined in Equation (\ref{eq:gradient}):

\begin{equation}
g_{x_t} = \left\| \frac{\partial J}{\partial x_t} \right\|_2, \quad 
g_{x_0} = \left\| \frac{\partial J}{\partial x_0} \right\|_2, \quad
r_{x_t} = \frac{g_{x_t}}{g_{x_t} + g_{x_0}}
\label{eq:gradient}
\end{equation}

Here, \(r_{x_t}\) quantifies the relative sensitivity of the learned Jacobian field to the current state \(x_t\) versus the start pose \(x_0\). A higher \(r_{x_t}\) indicates that the 
model's output depends less on the initial anchor—i.e., reduced start-pose bias in the learned representation.

With \(L_{\mathrm{cons}}\), the normalized error distribution shifts upward and becomes more concentrated, indicating reduced start-pose bias. The heatmap on the right 
further shows that temporal responses become more uniform along the time axis, confirming that \(L_{\mathrm{cons}}\) improves the modeling of mid-sequence high-frequency variations.
For the same interpolated state, a locally valid Jacobian should depend primarily on the current configuration neighborhood rather than on which endpoint is used as the anchor condition.
\begin{figure}[!ht]
    \centering
    \includegraphics[width=1\textwidth, trim=0 0 0 0, clip]{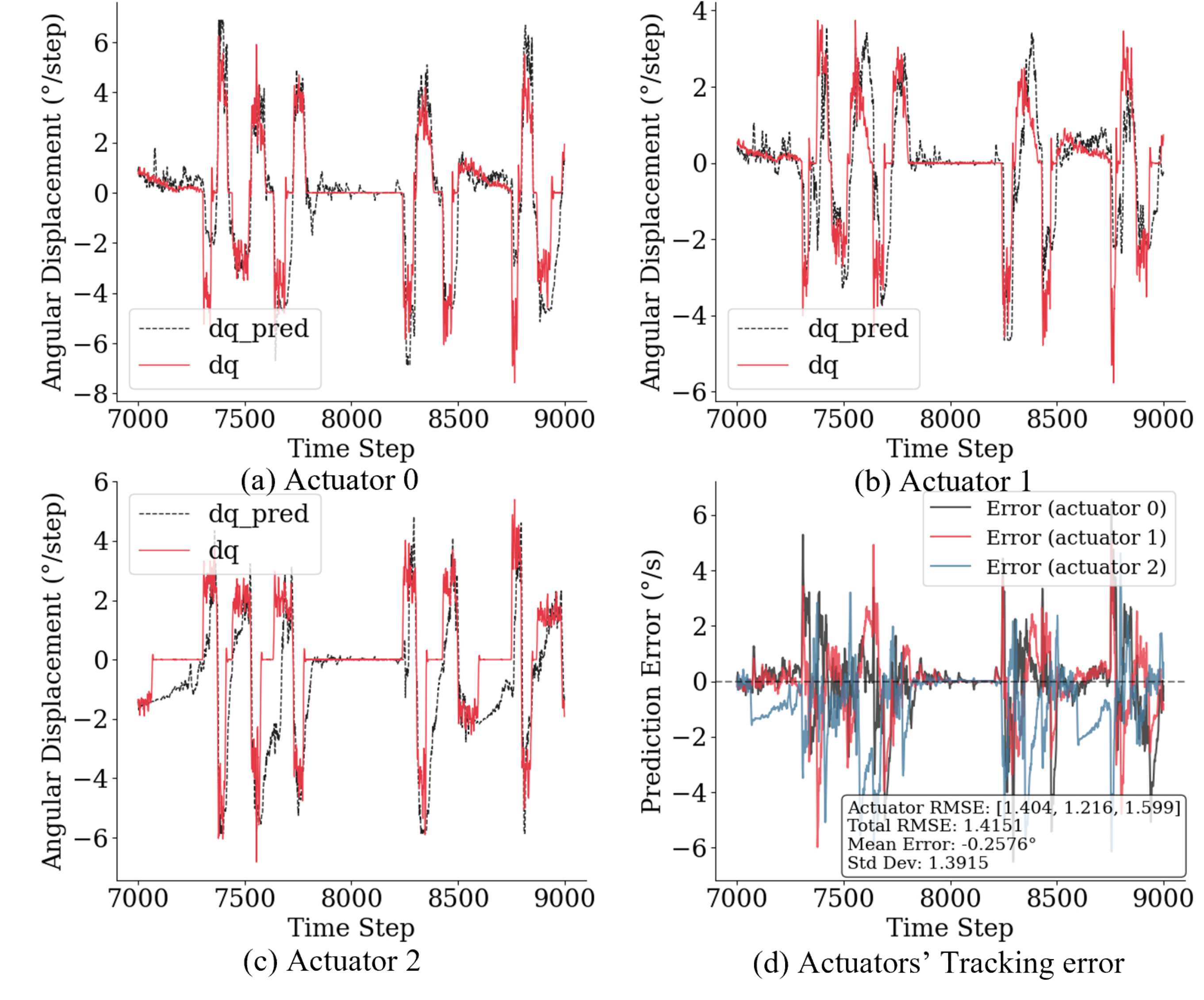}
    \caption{the inverse prediction results}
    \label{fig:inverse}
\end{figure}
\subsection{Open-Loop Inverse Validation}
\label{sec:4_4} 
To test whether the learned Jacobian field can recover feasible actuation commands from desired targets, we formulate the single-step planning problem in Equation (\ref{eq:inverse_opt}).

The optimization is performed without closed-loop feedback. As shown in Fig.~\ref{fig:inverse}(a-c), the predicted actuator commands closely match the real executed commands for 
most of the trajectory, confirming that the field is compatible with optimization-based planning. 

A systematic deviation is observed when the executed command abruptly drops to zero (e.g., at hard stops). In the training data, such hard stops are generated by directly commanding 
zero to the servo motors. The model, however, is not explicitly informed of these hard boundary conditions; it learns only from observed continuous motion and therefore produces a 
smoothly decaying prediction rather than an instantaneous jump. This behavior is a natural consequence of the smoothness assumption inherent to the continuous-time JFM formulation.

Quantitatively, as shown in Fig.~\ref{fig:inverse}(d), the overall RMSE is 1.4151, with a mean error of -0.2576° and a standard deviation of 1.39. These results further demonstrate 
the feasibility of using the learned Jacobian field for optimization-based planning.

The pointwise inference mode runs at approximately 3.5 ms per query, demonstrating feasibility for real-time deployment; full closed-loop integration with ODE-based inference is left 
for future work.

\section{Conclusion}
In this work, we address the nonlinear modeling challenges of rigid-flexible coupled bio-inspired dexterous fingers and the inherent contradiction between infinitesimal Jacobian models 
and discrete-time finite-step control. By leveraging Conditional Flow Matching (CFM), we propose a structured learning framework—Jacobian Flow Matching (JFM)—that learns a resolution-consistent 
Jacobian field, interpreting the actuation-motion relationship as a continuous velocity field consistent with observed finite displacements.

Experimental results on an anthropomorphic soft finger demonstrate that JFM significantly improves prediction accuracy and robustness. In single-step prediction, JFM reduces the global average 
RMSE by over 53\% compared to the pointwise baseline. Kernel density estimation shows that JFM predictions are highly concentrated in the low-error region (RMSE < 0.005), with density 
approximately 1.4\(\times\) that of the baseline, while successfully truncating the long tail of outliers. In multi-step open-loop prediction, ODE-based inference gradually outperforms 
pointwise inference as the stride increases. Under sparse sampling (stride=8), ODE inference improves the RMSE median by 14.43\% and reduces variance by 24.87\%. Inverse validation further 
confirms that the learned Jacobian field can recover feasible actuation commands from desired targets, with an overall RMSE of 1.4151.

These results demonstrate that JFM provides a reliable local kinematic model for rigid-soft coupled fingers, supporting both forward prediction and optimization-based planning.

While this work focuses on tendon-driven rigid-flexible coupled fingers, we speculate that the proposed framework may generalize to other dexterous hand systems that face similar challenges. Many 
soft and hybrid hands exhibit nonlinear actuation-motion mappings due to flexible materials (e.g., silicone-based soft hands, pneumatically driven actuators, or continuum structures), and their 
first-order kinematic relationships are difficult to derive analytically. Since JFM learns a continuous velocity field directly from observed trajectory data without requiring explicit physical 
parameters or material models, it is plausible that the same approach could be applied to other hand systems where paired actuation-motion trajectory data is available. However, direct validation 
on such systems remains an open question and is left for future work.

Future work will explore extending the framework to multi-finger coordination and out-of-plane motions, as well as integrating ODE-based inference into real-time control loops 
through model acceleration and distillation techniques.

\section*{Acknowledgments}
This work was supported by the National Natural Science Foundation of China (Grant No. 52305043 and No. 52175069).

\bibliographystyle{elsarticle-num}
\bibliography{jacobian_field_manu_2_0}

\end{document}


\section*{Supplementary Material}
\addcontentsline{toc}{section}{Supplementary Material}
\label{sec:supplementary}

\setcounter{section}{0}

\section{Notation}

\begin{description}
\item[$t$:] discrete sampling index.
\item[$x_t$:] the orientation of the phalanx at time index $t$. In this paper, we use Euler angles (rad) to represent it.
\item[$\Delta x_t$:] $x_{t+1} - x_t$.
\item[$q_t$:] the servo position (rad).
\item[$\Delta q_t$:] $q_{t+1} - q_t$.
\item[$J$:] the differential Jacobian matrix.
\item[$s_t$:] the observation/state feature vector available to the model/controller. In this model, $s_\tau = [x_\tau^T, x_t^T, q_t^T]^T$, $s_\tau = s_t + \tau \Delta s_t$, $\tau \in [0,1]$.
\end{description}

\section{Network Architecture \& Training Details}
\label{sec:network_arch_train}

\subsection{Jacobian Network Architecture}

\noindent
\textbf{Model overview:} We investigate two V-Net architectures for flow matching: 
a tiny-Transformer and an LSTM. Both networks are designed to predict 
per-phalanx outputs. They take an input tensor $x \in \mathbb{R}^{B \times J \times d_{\mathrm{in}}}$ 
and output $y \in \mathbb{R}^{B \times J \times d_{\mathrm{out}}}$, where $B$ is the 
batch size and $J$ is the number of joints (phalanx) modeled per sample. In this 
paper, we model a single finger with three phalanges (three joints). Each sample 
is represented as $J=3$ joint tokens (one token per phalanx/joint) and the network 
predicts per-joint outputs for this finger only.

\noindent\textbf{Input/Output:}
The model input dimension is $d_{\mathrm{in}} = 7$ $(x_\tau, x_t, q_t)$. 
The output dimension is $(J \times 3, \text{servo\_num})$.

\noindent\textbf{Embedding and conditioning:}
Each joint token is first projected to width $W = 64$ using a linear layer. 
Optionally, we add a learnable joint-type embedding $e(j) \in \mathbb{R}^W$ if joint 
identifiers are provided. Specifically, given \texttt{num\_joint\_types} = 32, an 
embedding table $\mathrm{Embedding}(32, W)$ is added token-wise:
\[
z \gets z + e(\mathrm{finger\_joint\_ids})
\]
where $\mathrm{finger\_joint\_ids}$ has shape $[B, J]$. (If joint types are not used, 
this term is omitted.) We optionally support sinusoidal positional encoding 
(disabled by default). When enabled, we add a fixed sinusoidal encoding to joint tokens.

\noindent\textbf{Transformer backbone:}
The encoder contains depth stacked Transformer blocks (default depth = 2). 
Each block applies:\\

\noindent
\textsl{Pre-norm multi-head self-attention:}  \\
$\mathrm{LayerNorm}(W) \rightarrow \mathrm{MultiheadAttention}(W, \text{num\_heads})$ \texttt{batch\_first=True} \\

\noindent
\textsl{Pre-norm feed-forward network (FFN):} 
\begin{align*}
\mathrm{Linear}(W, \text{ff\_mult} \cdot W) &\rightarrow \mathrm{GELU} \\
&\rightarrow \mathrm{Dropout} \\
&\rightarrow \mathrm{Linear}(\text{ff\_mult} \cdot W, W),
\end{align*}
with residual connections around both attention and FFN sublayers.

In our configuration, we use:
\begin{itemize}
\item width $W = 64$;
\item number of attention heads $H = 16$ (\texttt{num\_heads}=16);
\item FFN expansion $\text{ff\_mult} = 2$;
\item dropout $= 0.0$.
\end{itemize}

\noindent
\textsl{Output head:}
A final $\mathrm{LayerNorm}(W)$ followed by a linear projection produces per-token 
outputs:
\[
y = \mathrm{Linear}(W, d_{\mathrm{out}})(\mathrm{LayerNorm}(z))
\]

\noindent\textbf{LSTM backbone:}
The encoder contains stacked LSTM layers (default \texttt{num\_layers}=2). 
Each layer applies:

\noindent
\textsl{Input projection:}
\[
z = \mathrm{Linear}(d_{\mathrm{in}}, W)(x)
\]

\noindent
\textsl{Joint-type embedding (optional):}
\[
z \gets z + \mathrm{Embedding}(\text{num\_joint\_types}, W)(\mathrm{finger\_joint\_ids})
\]

\noindent
\textsl{LSTM encoder:}
\[
\text{out}, (h_n, c_n) = \mathrm{LSTM}(z), \quad \texttt{batch\_first=True}
\]

\noindent
\textsl{Output head:}
\[
\begin{aligned}
\text{out} &= \mathrm{LayerNorm}(W)(\text{out}) \\
y &= \mathrm{Linear}(W, 9)(\text{out}) \\
y &= \mathrm{reshape}(y, (B, J \times 3, 3))
\end{aligned}
\]

In our configuration, we use:
\begin{itemize}
\item input dimension $d_{\mathrm{in}} = 7$ $(x_\tau, x_t, q_t)$;
\item hidden dimension $W = 128$;
\item number of LSTM layers $= 2$;
\item number of joint types $= 32$;
\item output dimension per joint $= 9$ (reshaped to $3 \times 3$);
\item dropout $= 0.0$.
\end{itemize}

\subsection{Training Details}

We train the network using a single set of hyperparameters across all experiments (unless otherwise stated).

\noindent
\textbf{Randomness:}
\[
\text{random seed} = 42
\]

\noindent
\textbf{Optimizer:}
\[
\text{AdamW}, \quad \text{lr} = 1 \times 10^{-4}, \quad \text{weight decay} = 1 \times 10^{-2}, \quad \beta = (0.9, 0.95)
\]
Gradient accumulation is disabled (\texttt{gradient\_accumulate\_every}=1).

\noindent
\textbf{Learning-rate schedule:}
\[
\text{cosine decay with linear warm-up of } 3000 \text{ steps}
\]

\noindent
\textbf{Training length and evaluation cadence:}
\begin{itemize}
\item Train for up to $800$ epochs, with a maximum of $10^6$ training steps;
\item Validation run every $50$ epochs;
\item Checkpoints saved every $50$ epochs;
\item Rollouts generated every $10$ epochs.
\end{itemize}

\noindent
\textbf{Data split:}
\[
\text{train/val/test} = 80\% / 10\% / 10\%
\]

\noindent
\textbf{Data loading:}
\begin{itemize}
\item Batch size: train $= 256$, validation $= 128$, test $= 32$;
\item Number of workers: train $= 16$, validation/test $= 8$;
\item Pinned memory and persistent workers enabled.
\end{itemize}

\noindent
\textbf{Learning rate schedule details:}
The learning rate is updated per training step using a linear warm-up plus cosine annealing schedule. Specifically, for a total of $T = \text{max\_train\_steps}$ steps (default $10^6$), we warm up for $T_w = \text{lr\_warmup\_steps}$ steps ($3000$) from $10^{-3} \cdot \text{lr}$ to $\text{lr}$, followed by cosine annealing for the remaining steps with $\eta_{\min} = 5 \times 10^{-6}$.

\section{Experimental Platform \& Data Acquisition}

\subsection{Data Collection Pipeline}

\begin{figure}[H]
\centering
\includegraphics[width=1\textwidth, trim=0 0 0 0, clip]{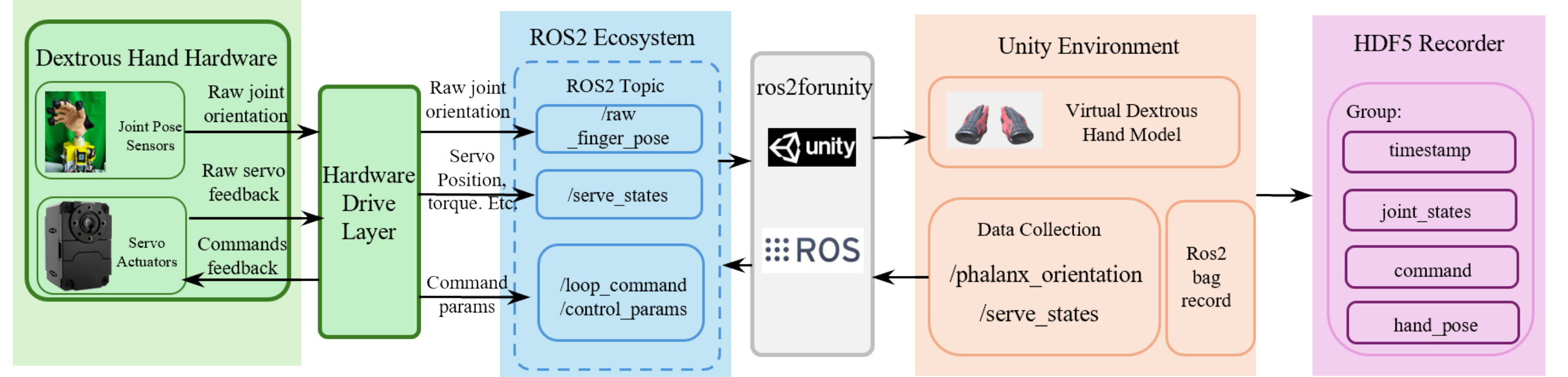}
\caption{Data collection pipeline overview}
\label{fig:data_collector}
\end{figure}

Figure~\ref{fig:data_collector} summarizes our data-collection pipeline. 
A dexterous finger hardware platform provides raw joint-pose sensing and low-level 
servo feedback, which are interfaced through a hardware driver layer and exposed 
in a ROS2 ecosystem.

During data collection, we log the following ROS2 topics:
\begin{itemize}
\item \texttt{/raw\_finger\_pose}: phalanx orientation;
\item \texttt{/servo\_states}: servo feedback;
\item control signals (e.g., loop commands and associated control parameters).
\end{itemize}

In parallel, we use a Unity-based virtual hand environment connected via 
\texttt{ros2-for-unity} to facilitate synchronized recording and visualization. 
All streams are recorded as ROS2 bags and exported into an HDF5 dataset, organized 
into groups including \texttt{timestamp}, \texttt{joint\_states}, \texttt{command}, 
and \texttt{hand\_pose}, enabling consistent offline training and evaluation.

\subsection{Data Specifications}

We collect on the order of $10^4$ samples. The data streams are recorded at 
different frequencies:
\begin{itemize}
\item Unity hand pose stream (\texttt{hand\_pose}): $45$ Hz;
\item Servo state stream (\texttt{servo\_states}): $55$ Hz.
\end{itemize}

For learning, we temporally align streams using timestamps (nearest-neighbor 
resampling to the target rate) before exporting ROS2 bags to an HDF5 dataset.

\subsection{Data Acquisition Configuration Summary}

\begin{table}[H]
\centering
\caption{Data acquisition parameters}
\label{tab:data_acquisition}
\begin{tabular}{l c}
\hline
\textbf{Parameter} & \textbf{Value} \\
\hline
Total samples & $\sim 10^4$ \\
Hand pose frequency & $45$ Hz \\
Servo state frequency & $55$ Hz \\
Resampling method & Nearest-neighbor \\
Output format & HDF5 \\
\hline
\end{tabular}
\end{table}